\documentclass[letterpaper, 10 pt, conference]{ieeeconf-2}  

\IEEEoverridecommandlockouts                              
\usepackage{amsmath}
\usepackage{amssymb}  
\usepackage{color}
 \usepackage{dsfont}
 \usepackage{cite}
 \usepackage{url}
 \usepackage{graphicx}
 \newtheorem{theorem}{Theorem}
\newtheorem{lemma}{Lemma}
\newtheorem{remark}{Remark}

\newtheorem{corollary}{Corollary}
\newtheorem{proposition}{Proposition}
\newtheorem{definition}{Definition}

\title{\LARGE \bf
Universal Approximation Property of Hamiltonian Deep Neural Networks
}

\def\Real{\mathbb{R}}
\def\Rn{\mathbb{R}^n}
\def\Rtn{\mathbb{R}^{2n}}
\def\Rp{\mathbb{R}^p}

\def\Realp{\mathbb{R}_+}

\def\Nat{\mathbb{N}}
\def\C{\mathcal{C}}
\def\P{\mathcal{P}}
\def\Wt{\tilde{W}}
\def\bt{\tilde{b}}
\def\dt{\tilde{d}}
\def\rt{\tilde{r}}
\def\etat{\tilde{\eta}}
\def\gammat{\tilde{\gamma}}
\def\At{\tilde{A}}

\def\L{\mathcal{L}}

\def\ntn{{n\times n}}

\def\e{\textnormal{e}}

\author{Muhammad Zakwan, Massimiliano d'Angelo and Giancarlo Ferrari-Trecate 
\thanks{This research is supported by the Swiss National Science Foundation under the NCCR Automation (grant agreement 51NF40 180545).}
\thanks{Muhammad Zakwan and Giancarlo Ferrari Trecate are with the Institute of Mechanical Engineering, Ecole Polytechnique Fédérale de Lausanne (EPFL), CH-1015 Lausanne, Switzerland {\tt\{muhammad.zakwan, giancarlo.ferraritrecate\}@epfl.ch}.  Massimiliano d'Angelo is with Sapienza University of Rome, Rome, 00185, Italy and with the Istituto di Analisi dei Sistemi e Informatica, Italian National Research Council (IASI-CNR), Rome, 00185, Italy, {\tt mdangelo@diag.uniroma1.it}. Corresponding author: M. d'Angelo.}%
}
\begin{document}

\maketitle
\thispagestyle{empty}
\pagestyle{empty}

\begin{abstract}
This paper investigates the universal approximation capabilities of Hamiltonian Deep Neural Networks (HDNNs) that arise from the discretization of Hamiltonian Neural Ordinary Differential Equations. 
Recently, it has been shown that HDNNs enjoy, by design, non-vanishing gradients, which provide numerical stability during training. However, although HDNNs have demonstrated state-of-the-art performance in several applications, a comprehensive study to quantify their expressivity is missing. In this regard, we provide a universal approximation theorem for HDNNs and prove that a portion of the flow of HDNNs can approximate arbitrary well any continuous function over a compact domain. This result provides a solid theoretical foundation for the practical use of HDNNs.
\end{abstract}

\section{INTRODUCTION}
Deep Neural Networks (DNNs) have been crucial for the success of machine learning in several real-world applications like computer vision, natural language processing, and reinforcement learning.
To achieve state-of-the-art performance, a common approach in machine learning is to increase the Neural Network (NN) depth. For instance, Convolutional Neural Networks (CNNs) AlexNet \cite{krizhevsky2017imagenet}, Visual Geometric Group (VGG) network, GoogLeNet/Inception \cite{szegedy2015going}, Residual Network (ResNet) \cite{he2016deep}, or recently developed transformers such as ChatGPT, contain hundreds to thousands of layers. 
It has been empirically demonstrated that deeper networks yield better performance than single-hidden-layer NNs for large-scale and high-dimensional problems \cite{shen2021neural, lin2018resnet}. 
However, a rigorous characterization of the approximation capabilities of complex NNs is often missing. Moreover, the understanding of how NN architectures (depth, width, and type of activation function) achieve their empirical success is an open research problem \cite{raghu2017expressive}. 

To quantify the representational power of NNs, researchers have focused on studying their Universal Approximation Properties (UAPs), namely their ability to approximate any desired continuous function with an arbitrary accuracy. 
To this aim,  several UAP results for various classes of NNs have been proposed. The UAP of Shallow NNs (SNNs), \emph{i.e.} with single hidden layer has proven in the seminal works of  Cybenko \cite{cybenko1989approximation} and Hornik \cite{hornik1989multilayer}. Exploiting the latter arguments, researchers have provided several results on UAPs for DNNs. 
For instance, in \cite{shen2021neural} it is proved that a DNN with three hidden layers and specific types of activation functions has the UAP. 
The paper \cite{lin2018resnet} demonstrates that a very deep ResNet, with stacked modules having one neuron per hidden layer and rectified linear unit (ReLU) activation functions, can uniformly approximate any integrable function. However, extending these results to other classes of activation functions is not straightforward.  We defer the interested readers to \cite{guhring2020expressivity} for a detailed survey on the subject.

Recently, an alternate representation of DNNs as dynamical systems has been proposed \cite{haber2017stable}. This idea was later popularized as Neural Ordinary Differential Equations (NODEs) \cite{chen2018neural}. By viewing DNNs through a dynamical perspective, researchers have been able to utilize tools from system theory in order to analyze their properties (\emph{e.g.}, Lyapunov stability, contraction theory, and symplectic properties).
Similar to DNNs, there are some contributions on UAPs for NODEs.
It has been shown in \cite{zhang2020approximation} that capping a NODE with a single linear layer is sufficient to guarantee the UAP, but exclusively for non-invertible continuous functions.
Furthermore, in \cite{tabuada2022universal}, differential geometric tools for controllability analysis were used to provide UAPs for a class of NODEs, while in \cite{li2022deep}, the compositional properties of the flows of NODEs were exploited to obtain UAPs. In \cite{tabuada2022universal} certain restrictions on the choice of activation functions are present, whereas \cite{li2022deep} impose constraints on the desired target function. 
Finally in \cite{celledoni2022dynamical}, some interesting tools, such as composition of contractive, expansive, and sphere-preserving flow maps, have been used to prove a universal approximation theorem for the flows of dynamical systems. 

Although DNNs tend to empirically perform well in general, the increasing depth can also present challenges, such as the vanishing/exploding gradient problem during the training via gradient descent algorithms. 
These phenomenon happen when the gradients computed during back-propagation either approach to zero or diverge. 
In such cases, the learning process may stop prematurely or become unstable, thereby limiting the depth of DNNs that can be utilized and consequently preventing the practical exploitation of UAP in DNNs.
Practitioners have proposed several remedies to address these challenges, including skip connections in ResNet \cite{he2016deep}, batch normalization, subtle weights initialization, regularization techniques such as dropout or weight decay, and gradient clipping \cite{goodfellow2016deep}. 
However, all of these ad hoc methods do not come with provable formal guarantees of non-vanishing gradients. Recently, a class of DNNs called Hamiltonian Deep Neural Networks (HDNNs) have been proposed in \cite{galimberti2023hamiltonian}. These DNNs stem from the discretization of Hamiltonian NODEs, and enjoy non-vanishing gradients \emph{by design} if {symplectic} discretization methods \cite{hairer2006geometric} are used \cite{galimberti2023hamiltonian}. Moreover, the expressivity of HDNNs has been demonstrated empirically on several benchmarks in classification tasks. Nevertheless, the theoretical foundation on the UAP of HDNNs has yet to be explored.

\subsection{Contributions}
In this paper, we present a rigorous theoretical framework to prove a UAP of HDNNs. First, with a slight modification, we generalize the class of HDNNs considered in \cite{galimberti2023hamiltonian} without compromising the provable non-vanishing gradients property\footnote{For the sake of simplicity, we retain the same name and also refer to the proposed modified version as HDNNs.}. 
Second, we prove that a portion of the flow of HDNNs can approximate any continuous function with arbitrary accuracy.   To the best of our knowledge, this is the first UAP result for a class of ResNets enjoying non-vanishing gradients which are essential for numerically well-posed training.   The proof is based on three essential features \emph{i.e.} symplectic discretization through the Semi-Implicit Euler (SIE) method, a careful choice of initial conditions, and an appropriate selection of the flow. 
It is important to note that general DNNs, such as deep Multi-Layered Perceptrons (MLPs) or recurrent NNs, can suffer from vanishing gradients and might fail to approximate arbitrary functions if the training stops early. 
Third, since DNNs arising from the discretization of ODEs are automorphic maps -- they do not alter the dimension of the input data -- based on the composition of functions, we
extend the main result to approximate maps, where the dimensions of domain and co-domain are different. 
Finally, we provide a characterization of the approximation error with respect to the depth.

{\it Organization:} Section \ref{sec:preliminaries} provides preliminaries on Hamiltonian NODEs, the employed discretization scheme, definitions of UAPs, and the problem formulation. In Section \ref{sec:main}, we prove the UAP for HDNNs (Theorem \ref{th:UAP_hamiltonian}), we investigate the case when the desired function is not an automorphic map (Corollary \ref{cor:problem2}), and provide some remarks on the approximation error (Proposition \ref{prop:approx}). We discuss a numerical example in Section \ref{Sec:numerical_examples}.  Finally, conclusions are drawn in Section \ref{sec:conslusions}.  
\subsection{Notation}
We denote the set of non-negative reals with $\Real_+$. For a vector $x \in \mathbb{R}^n$, its 2-norm is represented by $\| x \|$ and its 1-norm $\|x\|_1 := \sum_j |x_j|$. Given an $\L_2$-function $f:\Rn \rightarrow \Rn$ the $\L_2$ norm over the compact set $\Omega \subset \Real^n$ is denoted by $\|f\|_{\L_2(\Omega)}$ and the (essential) supremum norm by $\|f\|_{\L_\infty(\Omega)} = \sup_{x\in\Omega} \|f(x)\|$.  $|A|$ stands for the determinant of a squared matrix $A$. We represent with $0_n$ the zero vector in $\Rn$ and with $0_\ntn$ the matrix with all entries equal to zero in $\Real^\ntn$. We denote the column vector of ones of dimension $n$ with $\mathds{1}_n$. Given $\Omega \subset \Real^n$, $\C(\Omega;\Real^n)$ stands for the space of continuous functions $f: \Omega \to \Rn$. Given $T\in \Real_+$, we refer to $\P([0,T];\Real^p)$ as the space of piecewise constant function $\theta : [0,T] \to \Real^p$. Functions that cannot be represented in the form of a polynomial are referred to as \emph{non-polynomial} functions.
    
\section{Preliminaries and problem formulation}\label{sec:preliminaries}
\subsection{Hamiltonian Neural Ordinary Differential Equations}\label{subsec:HNODE}
A Neural ODE \cite{chen2018neural} (NODE) is represented by the dynamical system for $t\in [0, T]$ given by
\begin{align}
  \label{eq:NODE}
  \dot{x}(t)=F(x(t), \theta(t)) \quad \textnormal{with} \ x(0)= x_0 \in \Omega \;,
\end{align}
where $x(t)\in \mathbb{R}^n$ is the state at time $t$ of the NODE with initial condition $x_0$ in some compact set $\Omega \in \Rn$, and $F: \Rn \times \Rp \to \Rn$ is such that $F(x, \theta)$ is Lipschitz continuous with respect to $x$ and measurable with respect to the weights $\theta$. We further assume that $\theta(t) \in \P([0,T];\Real^p)$.
When used in machine learning tasks, the NODE is usually pre- and post- pended with additional layers, \emph{e.g.,} $x_0 = \varphi_{\alpha}(z)$, with $z \in \mathbb{R}^{n_z}$ the input and $\varphi_{\alpha}$ a NNs with parameters $\alpha\in \mathbb{R}^{n_\alpha}$, and the output $y$ is computed as $y=\phi_{\beta}(x(T))$, where $\phi_{\beta}$ is a NNs with parameters $\beta\in \mathbb{R}^{n_\beta}$.

In this paper, we consider a class of NODEs inspired by Hamiltonian systems. In particular, we consider the Hamiltonian function $H: \Rtn \times \Realp \to \Real$ given by 
\begin{equation}\label{eq:Hamiltonian}
    H(x,t) = \tilde{\sigma}( W(t)x + b(t))^\top \mathds{1}_n + \eta(t)^\top\, x\;,
\end{equation}
where $W: \Realp \to \Real^{2n\times 2n}$, $b: \Realp \to \Rtn$, $\eta : \Realp \to \Rtn$ are piece-wise constant, while $\tilde{\sigma} : \Real \to \Real$  is a differentiable map, applied element-wise when the argument is a matrix, and such that $\sigma(x):= \frac{\partial \tilde\sigma}{\partial x}(x)$ is non-polynomial and Lipschitz continuous. As explained below, $\sigma$ will play the role of the so-called \emph{activation function}.
Examples that satisfy the above assumptions are provided in Table \ref{tab:act}. Note that if we set $\eta(t) = 0$ in \eqref{eq:Hamiltonian}, we recover DNNs proposed in \cite{haber2017stable,galimberti2023hamiltonian}.
We define the Hamiltonian system 
\begin{align}\label{eq:Hamiltonian_dynamics_general}
  \dot{x}(t)=J(t) \frac{\partial H (x(t),t)}{\partial x} \;,
\end{align}
where $J(t)$ is piecewise constant skew-symmetric matrix, namely $J(t)=-J(t)^\top$, in $\Rtn\times \Rtn$ for any $t\geq 0$. By taking into account the expression of the Hamiltonian in \eqref{eq:Hamiltonian}, we obtain the following dynamics
\begin{align}\label{eq:Hamiltonian_dynamics}
  \dot{x}(t)=J(t)\left(W(t)^\top \sigma\big(W(t)x(t) + b(t)\big)+ \eta(t)\right) \;.
\end{align}
Note that the latter equation can be written in the form \eqref{eq:NODE}, when the weights are given by $\theta(t) = \{J(t), W(t),b(t),\eta(t)\}$ for $t\in [0,T]$.\\
For the numerical implementation of NODE \eqref{eq:Hamiltonian_dynamics}, we rely on the SIE discretization \cite{hairer2006geometric} because it can preserve the symplectic flow of time-invariant Hamiltonian systems and is crucial to prove non-vanishing gradient property of the resulting HDNNs (further details will be given in the next section).
In particular, splitting the state of the Hamiltonian systems into $x=(p,q)$, we obtain the HDNN 
\begin{equation}
\label{eq:Hamiltonian_dynamics_general_SIE}
\begin{aligned}
{\left[\begin{array}{c}
{p}_{j+1} \\
{q}_{j+1}
\end{array}\right] } & =\left[\begin{array}{c}
{p}_j \\
{q}_j
\end{array}\right]+ h J_j\left[\begin{array}{c}
\frac{\partial H}{\partial {p}}\left({p}_{j+1}, {q}_j, t_j\right) \\
\frac{\partial H}{\partial {q}}\left({p}_{j+1}, {q}_j, t_j\right)
\end{array}\right],
\end{aligned}
\end{equation}
where $h = T\slash N$, with $N \in \Nat$, is the integration step-size, $j=0,\dots,N-1$ and $p_j$ and $q_j$ are the two state components in $\Real^{n}$. Moreover, by taking into account the expression of the Hamiltonian in \eqref{eq:Hamiltonian}, namely the dynamics \eqref{eq:Hamiltonian_dynamics}, we obtain the following difference equation
\begin{equation}
\label{eq:Hamiltonian_dynamics_SIE}
\begin{aligned}
& {\left[\begin{array}{c}
{p}_{j+1} \\
{q}_{j+1}
\end{array}\right] }  =  \left[\begin{array}{c}
{p}_j \\
{q}_j
\end{array}\right] \\ 
&\qquad + h J_j \left( W_j^{\top} \sigma\left({W}_j\left[\begin{array}{c}
{p}_{j+1} \\
{q}_j
\end{array}\right]+{b}_j\right)  +  \eta_j \right).
\end{aligned}
\end{equation}
Clearly, the set of weights is given by $\theta_j = \{J_j, W_j,b_j,\eta_j\}$ with $j=0,\dots,N-1$. With a little abuse of notation we write $\theta_j \in \Rp$ with $j=0,\dots,N-1$ and appropriate $p\in \Nat$.
Although, in general, one has to compute the update $(p_{j+1},q_{j+1})$ of \eqref{eq:Hamiltonian_dynamics_SIE} through an implicit expression, it is possible to rewrite it in an explicit form, when the matrices $J_j$ and $W_j$ satisfy some assumptions, \emph{e.g.}, by choosing $J_j$ block anti-diagonal and $W_j$ block diagonal~\cite{galimberti2023hamiltonian}. 

\subsection{Universal Approximation Property}
In this section, we present some essential definitions pertaining to universal approximation properties.
\begin{definition} [UAP of a function]\label{def:UAPmaps}
Consider a function $g_\theta: \Real^n \to \Real^n$ with parameters $\theta\in \Rp$ and a compact subset $\Omega \in \Real^n$, then $g_\theta$ has the Universal Approximation Property (UAP) on $\Omega\subset \Real^n$ if for any $f\in \C(\Omega;\Real^n)$ and $\varepsilon>0$, there exists $\theta \in \Rp$ such that
\begin{equation} \label{eq:UAP_scalar}
\sup_{x\in\Omega} \|f(x) - g_\theta(x)\| \leq \varepsilon \; .
\end{equation}
\end{definition}

We provide the following fact which descends from \cite{pinkus1999approximation}.
\begin{proposition}\label{prop:UAPmaps}
Let $\sigma\in \C(\Real;\Real)$ be non-polynomial, then for any $f\in \C(\Omega;\Rn)$, where $\Omega \in \Real^n$, and $\varepsilon>0$, there exist $N\in\Nat$, $A_j,W_j \in \Real^{n \times n}$ and  $b_j \in \Rn$ such that the function $g:\Real^n \to \Rn$ given by
\begin{equation}\label{eq:Cybenko_sum_matrix}
    g(x):= \sum_{j = 0}^{N-1} A_j \,\sigma(W_j x + b_j) \;,
\end{equation} 
satisfies
\begin{equation} \label{eq:UAP_vector}
\sup_{x\in\Omega} \|f(x) - g(x)\| \leq \varepsilon \;.
\end{equation}
\end{proposition}
Some examples of activation functions $\sigma$, such that $g$ in  \eqref{eq:Cybenko_sum_matrix} satisfies the UAP, are given in Table \ref{tab:act}.
\begin{table}[]
    \centering
    \begin{tabular}{c|c}
    \hline
    Activation Function  & $\sigma(x)$ \\
    \hline 
     ReLU    &  $\max \{x,0 \}$\\
     Sigmoidal  & $(1 + \exp(-x))^{-1}$ \\
     Softplus    & $\log (1 + \exp(x)) $ \\
     Hyperbolic Tangent & $\tanh(x)$ \\
     Radial Basis Function & $\frac{1}{\sqrt{2 \pi}} \exp(-\frac{x^2}{2})$ \\
     \hline
    \end{tabular}
    \caption{Examples of activation functions.}
    \label{tab:act}
\vspace{-0.9cm}
\end{table}



In the sequel, we refer to the UAP with bound $\varepsilon>0$ to quantify the estimation error in equations \eqref{eq:UAP_scalar}, and \eqref{eq:UAP_vector}. This value is typically a function of $N$, $n$, and the desired $f$, and it is  characterized in Proposition \ref{prop:approx}.
\subsection{Problem formulation}\label{subsec:problem_formulation}
The goal of our paper can be formulated as follows.
{\it Problem 1: Let $f: \Rn \to \Rn$ be a continuous function, $\Omega \subset \Rn$ be a compact set, and $\varepsilon>0$ be the desired approximation accuracy. Find $N\in\Nat$ and weights $\theta_j = \{J_j, W_j,b_j,\eta_j\}$ with $j=0,\dots,N-1$ of \eqref{eq:Hamiltonian_dynamics_SIE}, such that a portion $\varphi : \Rn \to \Rn$ of the flow $\Phi_N : \Real^{2n} \to \Real^{2n}$ at time $N\in\Nat$ of  \eqref{eq:Hamiltonian_dynamics_SIE} has the UAP on $\Omega$.}\\
We recall that the flow at time $N\in\Nat$ of \eqref{eq:Hamiltonian_dynamics_SIE} is the corresponding unique solution at time $N\in\Nat$. In particular, $\Phi_N : \Real^{2n} \to \Real^{2n}$ is the flow at time $N$ as function of the initial condition. The flow $\varphi$ will be precisely defined in Theorem \ref{th:UAP_hamiltonian}.

Moreover, motivated by real-world applications, we are also interested in approximating arbitrary continuous functions $f: \mathbb{R}^n \rightarrow \mathbb{R}^r$ where $r$ is not necessarily equal to $n$. For instance, in classification tasks, typically $r < n$, as $r$ corresponds to the number of classes to be classified and $n$ represents the number of features.
We address this problem in Corollary \ref{cor:problem2}.

\section{Main Results}\label{sec:main}
In this section, we present our main results whose proofs are given the Appendix.
We address the Problem 1 in Theorem \ref{th:UAP_hamiltonian}, which is a universal approximation theorem for the HDNN \eqref{eq:Hamiltonian_dynamics_general_SIE}. 

\begin{theorem} \label{th:UAP_hamiltonian}
Consider the discrete-time system \eqref{eq:Hamiltonian_dynamics_SIE} with initial condition $(p_0,q_0) = (\xi,0_n)$, for some $\xi \in \Omega$ with $\Omega \subset \Rn$ compact. Then, the \emph{restricted} flow $\varphi : \xi \mapsto q_N$ has the UAP on $\Omega$.
\end{theorem}

In other words, Theorem \eqref{th:UAP_hamiltonian} states that given  the system \eqref{eq:Hamiltonian_dynamics_SIE} with initial condition $(p_0,q_0) = (\xi,0_n)$, for any $f\in C(\Omega;\Real^n)$ and $\varepsilon>0$, there exist $N\in \Nat$ and weights $\theta_j = \{J_j, W_j,b_j,\eta_j\}$ with $j=0,\dots,N-1$ such that the function $\varphi: \xi \mapsto q_N$  satisfies
\begin{equation}\label{eq:app_xi}
    \sup_{\xi\in\Omega} \|f(\xi) - \varphi(\xi)\| < \varepsilon \;.
\end{equation}

\begin{remark}[Key ingredients for UAP]
The proof of Theorem \ref{th:UAP_hamiltonian}, besides exploiting arguments from \cite{cybenko1989approximation,hornik1989multilayer} for showing UAPs, it is based on three critical key steps: \emph{i)} the SIE discretization scheme, \emph{ii)} the initial condition $(p_0,q_0) = (\xi,0_n)$, and \emph{iii)} the focus on the \emph{restricted} flow $\xi \mapsto q_N$, which refers to map the initial condition of the $p$ state to the flow of the $q$ state.
\end{remark}

In particular, the choice of the SIE discretization scheme together with the initial condition $(p_0,q_0) = (\xi,0_n)$ allows one to exploit the framework of Cybenko \cite{cybenko1989approximation}  to express the function $\varphi : \xi \mapsto q_N$ as \eqref{eq:Cybenko_sum_matrix} (see equation \eqref{eq:def_varphi} in the Proof of Theorem \ref{th:UAP_hamiltonian} in Appendix \ref{app:th_main}).

\begin{remark}[Feature augmentation]
By defining the flow $\Phi_N$ of the discrete-time system \eqref{eq:Hamiltonian_dynamics_SIE} (evolving in $\Real^{2n}$), we note that \eqref{eq:app_xi} can be written as
\begin{equation} 
\sup_{x\in\Omega} \|f(x) - \pi \circ \Phi_N \circ \iota (x) \| \leq \varepsilon\;,
\end{equation}
where $\iota: \Real^{n} \to \Real^{2n}$ is the injection given by $\iota(z_1,\dots,z_n) = (z_1,\dots,z_n,0,\dots,0)$ and $\pi : \Real^{2n} \to \Real^n$ is the projection $\pi(x_1,\dots,x_n,x_{n+1},\dots, x_{2n}) = (x_{n+1},\dots,x_{2n})$.
This is equivalent to the common practice in machine learning of augmenting the size of the feature space \cite{dupont2019augmented}. 
It has been demonstrated that this technique can improve DNN performance in several learning tasks. 
Moreover, it is also closely related to the idea of extended space \cite{goodfellow2016deep}, which suggests that by increasing the dimensionality of the feature space, one can capture more complex relationships.
\end{remark}

We note that the UAP results in \cite{tabuada2022universal} do not apply in our framework because of the skew-symmetric matrix $J$ multiplying the partial derivative of the Hamiltonian in \eqref{eq:Hamiltonian_dynamics_general}. 
Moreover, we provide UAPs directly for implementable discrete-time layer equations \eqref{eq:Hamiltonian_dynamics_SIE} instead of the continuous-time NODEs. Indeed, an arbitrary discretization method may not conserve the desired properties, making it challenging to prove the UAP of discretized NODEs in general. 

Untill this point, we focused on automorphisms on $\mathbb{R}^n$. The next result presents the UAP of a general map from $\Omega \subset \Real^n$ to $\mathbb{R}^r$.

\begin{corollary}\label{cor:problem2}
Consider the discrete-time system \eqref{eq:Hamiltonian_dynamics_SIE} with initial condition $(p_0,q_0) = (\xi,0_n)$, for some $\xi \in \Omega$ with $\Omega \subset \Rn$ compact, and  the \emph{restricted} flow $\varphi : \xi \mapsto q_N$. 
Let $h: \Rn \to \Real^r$ be a Lipschitz continuous function such that $f\left(\Omega \right) \subseteq h\left(\mathbb{R}^n\right)$.
Then, for any $\varepsilon > 0$, the function $h\circ \varphi : \Omega \to \Real^r$, satisfies 
\begin{align}
   \sup_{\xi \in \Omega}\left\|f(\xi)- h \circ \varphi(\xi) \right\|\leq \varepsilon .
 \end{align}
\end{corollary}

A typical example that satisfies the necessity condition $f\left(\Omega \right) \subseteq h\left(\mathbb{R}^n\right)$ is $h(\varphi) = W_o^\top \varphi + b_o$, $W_o \in \Real^{n \times r}$, and $b_o \in \mathbb{R}^r$, which is common in classification problems. 
It is straightforward to see that $h(\cdot)$ is Lipschitz continuous, surjective, and satisfies the condition $f\left(\Omega\right) \subseteq h \left(\mathbb{R}^n\right)$. 

It is worth mentioning that unlike other papers \cite{shen2021neural,lin2018resnet,tabuada2022universal}, our results do not impose restrictive conditions on activation functions, which expands their potential applicability.

\subsection{Auxiliary properties of HDNNs}
In the following, we highlight a few associated properties of HDNNs. First, we provide a bound on the desired accuracy of the approximation error with respect to the depth of HDNNs. Second, we state a remark on their non-vanishing gradients property. 

Let us define the first absolute moment $C_f$ of the Fourier magnitude distribution of a desired function $f$. Thus, given $f : \Rn \to \Rn$, with a Fourier representation of the form $f(x) = \int_{\Rn} \e^{i \omega^\top x} \tilde{f}(\omega) \textnormal{d} x$, we define 
\begin{equation}\label{eq:Cf}
    C_f := \int_{\Rn} \|\omega\|_1 \|\tilde{f}(\omega)\| \textnormal{d}\omega \;.
\end{equation}
The condition \eqref{eq:Cf} is usually interpreted as the integrability of the Fourier transform of the gradient of the function $f$, and a vast list of examples for which bounds on $C_f$ can be obtained  are given in Section IX of \cite{barron1993universal}. 

\begin{proposition}\label{prop:approx}
    Consider the discrete-time system \eqref{eq:Hamiltonian_dynamics_SIE} with  sigmoidal\footnote{The function $\sigma(x)$ is assumed to be a sigmoidal function, if it is a bounded function on the real line satisfying $\sigma(x) \rightarrow 1$ as $x \rightarrow \infty$ and $\sigma(x) \rightarrow -1$ as $x \rightarrow -\infty$ \cite{barron1994approximation}.} $\sigma$ and initial condition $(p_0,q_0) = (\xi,0_n)$, for some $\xi \in \Omega=[-1,1]^n$. Then, the \emph{restricted} flow $\varphi : \xi \mapsto q_N$ has the UAP on $\Omega$ with bound $2^{\frac{n}{2}} \frac{C_f}{\sqrt{N}}$.
\end{proposition}

Proposition \ref{prop:approx} states that for any $f\in \mathcal{C}(\Omega;\Real^n)$ with finite $C_f$ and $N\in\Nat$, there exist  parameters $\theta_j = \{J_j, W_j,b_j,\eta_j\}$ with $j=0,\dots,N-1$, such that the function $\varphi : \xi \mapsto q_N$ satisfies
\begin{equation} 
\sup_{x\in\Omega} \|f(x) - \varphi(x)\| \leq 2^{\frac{n}{2}}\frac{C_f}{\sqrt{N}} \;.
\end{equation}
Further remarks on the evaluation\slash approximation of this bound can be found in \cite{barron1993universal} and \cite{barron1994approximation}.

As mentioned earlier, it has been shown that HDNNs considered in \cite{galimberti2023hamiltonian} are endowed with non-vanishing gradients  or in a special case, non-exploding gradients \cite{zakwan2022robust}, \emph{i.e.}, they ensure numerically well-posed training. We defer the reader to those papers for a formal discussion of the non-vanishing gradients property. 

\begin{remark}[Non-vanishing gradients]
The HDNN given by the discrete-time system \eqref{eq:Hamiltonian_dynamics_SIE} enjoys the non-vanishing gradients property when optimizing a generic loss function. In particular, this property is related to the Backward Sensitivity Matrix  $\frac{\partial x_N}{\partial x_{N-j}} = \prod_{\ell = N-j}^{N-1} \frac{\partial x_{\ell+1}}{\partial x_{\ell}}$, where $x =\left(p, q \right)$, at layer $N-j$ for $j=1,\dots, N-1$.
Although the considered Hamiltonian \eqref{eq:Hamiltonian} is different from the one of \cite{galimberti2023hamiltonian} (because of the linear term), one is able to prove the non-vanishing gradients property (by establishing a lower bound for the Backward Sensitivity Matrix) by following the same arguments of \cite[Theorem 2]{galimberti2023hamiltonian} which relies specifically on the symplectic property of the flow and not on the Hamiltonian structure.
\end{remark}

\section{Numerical Example} \label{Sec:numerical_examples}
In this example, our goal is to approximate the function $ y(x) = 2(2\cos(x)^2 - 1)^2 - 1$ considered in \cite{mhaskar2016learning}. The training set comprises  5000 datapoints generated by sampling $y(t)$ randomly for $x \in [-2\pi, 2\pi]$. We choose the mean square error as the loss function and compare the following NN architectures:

i) The SNN $\hat{y} = W_o \sigma (W x + b)$,
with  $W_o \in \mathbb{R}^{1 \times N_h}$, $W \in \mathbb{R}^{N_h \times 1}$ and $b \in \mathbb{R}^{N_h}$, where $N_h$ is the number of hidden neurons. We use the values of  $N_h$ in the set $\{ 400, 800, 1200, 1800, 2400 \}$.

ii) an HDNN, called HDNN-1, with forward equation \eqref{eq:Hamiltonian_dynamics_SIE} and weight matrices  \eqref{eq:parametrization_structure}
for $j = 0,1,\cdots,6$. 

iii) an HDNN, called HDNN-2, with forward equation \eqref{eq:Hamiltonian_dynamics_SIE}, where $W_j$ is block-diagonal for $j = 0,1,\cdots,3$ to match the number of parameters in HDNN-1. 

For HDNNs, we choose a sufficiently small step-size $h = 0.001$, and the initial conditions as $p_0,q_0 = ([x, 0_{M/2-1}], 0_{M/2})$, where $M$ is always an even integer. Moreover, the output equation is given by $\tilde{y} = W_o q_N + b_o$, where $W_o \in \mathbb{R}^{1 \times M/2}$ and $b_o \in \mathbb{R}^{M/2}$.
To have almost the same number of parameters in the chosen HDNNs, we choose $M$ from the set $\{24, 36, 44, 54, 62 \}$ for HDNN-1 and  $\{ 26,  36, 44, 54, 64\}$ for HDNN-2, respectively.

\begin{figure}
    \centering
    \includegraphics[width = \linewidth]{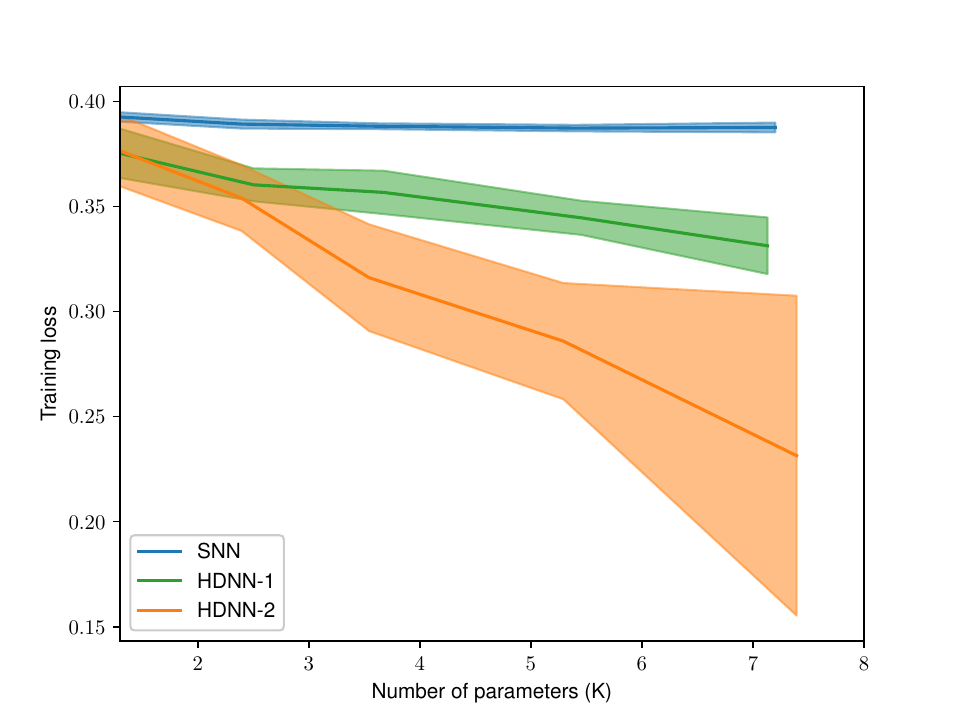}
    \caption{Averaged training loss and standard deviations over multiple experiments for three architectures.}
    \label{fig:my_label2}
    \vspace{-0.5cm}
\end{figure}

Fig.~\ref{fig:my_label2} shows that the training loss decreases when more parameters are used for all three architectures.
Moreover, we can see that for the same number of parameters, the block diagonal $W_j$ matrices of HDNN-2 with half the number of layers can be leveraged to further improve the performance over HDNN-1.

\section{Conclusion and Future Work}\label{sec:conslusions}
We demonstrated the universal approximation property of Hamiltonian Deep Neural Networks (HDNNs) that also enjoy non-vanishing gradients during training.  
This result affirms both the practicality and theoretical foundation of HDNNs. 
In particular, we have demonstrated that a portion of the flow of HDNNs can approximate any continuous function in a compact domain. 
Also, we provide some insights on the approximation error with respect to the depth of neural network. 

Our work opens doors to quantifying the expressivity of other physics-inspired neural networks with special properties, such as \cite{zakwan2022physically}.
Future research will focus on leveraging differential geometric tools \cite{ tabuada2022universal} to establish universal approximation properties for HDNNs, where the Hamiltonian function is parameterized by an arbitrary neural network. 
%

\appendix

\subsection{A preliminary lemma}\label{app:preliminary_lemma}

In order to prove Theorem \ref{th:UAP_hamiltonian}, we introduce a key auxiliary result which relaxes the necessity of full-rank weight matrices ${W}_j$ in \eqref{eq:Cybenko_sum_matrix} assumed in \cite[Theorem 2.6]{aizawa2020universal}. During the training of NN \eqref{eq:Cybenko_sum_matrix}, some entries of ${W}_j$ in \eqref{eq:Cybenko_sum_matrix} might vanish and this assumption cannot be satisfied. 
Therefore, the result in \cite{aizawa2020universal} might not be of practical use.
However, the following Lemma shows that even if $W_j$ in \eqref{eq:Cybenko_sum_matrix} are not full-rank, we can still construct an approximation with full-rank matrices and apply the results of \cite{cybenko1989approximation, pinkus1999approximation}.

\begin{lemma}
\label{lem:W_full_rank}
    Let $g$ be the function in \eqref{eq:Cybenko_sum_matrix} with the UAP on $\Omega$. For any $\tilde{\varepsilon}>0$ we can find $\At_j,\Wt_j \in \Real^{n \times n}$, with $\Wt_j$ full rank, and  $\bt_j \in \Rn$ for $j=0,\dots, N-1$, such that $\tilde g(x):= \sum_{j = 0}^{N} \At_j \,\sigma(\Wt_j  x + \bt_j)$ satisfies $\| \tilde{g} - g \|_{\L_\infty(\Omega)} \leq \tilde{\varepsilon}$.
\end{lemma}

In other words, Lemma \ref{lem:W_full_rank} allows us to assume, without loss of generality, that the function $g$ in \eqref{eq:Cybenko_sum_matrix} can be arbitrarily well-approximated by using full-rank matrices $W_j$ for any $j=0,\dots,N-1$.

\begin{proof}
  Given the function $g$ in \eqref{eq:Cybenko_sum_matrix} with the UAP on $\Omega$, we consider the case in which there exists the set $K = \{\kappa\in \{0,\dots,N-1\} :  |W_\kappa|=0\}$ non-empty with cardinality $\Tilde{n}$. For $\kappa \in K$, let $\textnormal{rank}(W_{\kappa}) = n-r_\kappa$, with $r_\kappa>0$ the number of dependent column vectors of $W_\kappa$ so that, up to a row permutation, assume $W_\kappa$ is partitioned as 
  \begin{equation}\label{eq:Wk_partion_proof}
  W_\kappa = \left(w_\kappa^{(1)},\dots,w_\kappa^{(r_\kappa)},w_\kappa^{(r_\kappa+1)},\dots,w_\kappa^{(n)} \right)^\top,
\end{equation}
    with the last $n-r_\kappa$ vectors linearly independent.
  Then, the parameters $\At_j,\Wt_j \in \Real^{n \times n}$ and  $\bt_j \in \Rn$ of the function $\tilde g(x)= \sum_{j = 0}^{N} \At_j \,\sigma(\Wt_j  x + \bt_j)$ can be selected as follows. We set $\At_j = A_j$, $\bt_j = b_j$ for all $j=0,\dots, N-1$. Moreover, $\Wt_j = W_j$ for all $j\notin K$ and, for $\kappa\in K$, $\Wt_\kappa = W_\kappa + \Lambda_\kappa$, where $\Lambda_\kappa = \left(\tilde{w}_\kappa^{(1)},\dots,\tilde{w}_\kappa^{(r_\kappa)},0_n,\dots,0_n \right)^\top$,
   and the vectors $\tilde{w}_\kappa^{(\ell)}$,  $\ell = 1,\dots,r_\kappa$, are selected such that $|\Wt_\kappa|\neq 0$ and 
\begin{align}\label{eq:ineq_wkappa_proof}
  \|\tilde{w}_\kappa^{(\ell)}\| & \leq \frac{\tilde \varepsilon}{r_\kappa \, \tilde{n} \, \sqrt{n} \,L_\sigma \, \|x\|_{\L_{\infty}(\Omega)}\,  \max_{1\leq p\leq n} \|a_\kappa^{(p)}\|}\;,
\end{align}
where $L_\sigma$ is the Lipschitz constant of function $\sigma$ and $a_{\kappa}^{(p)^\top}$, $p=1,\dots,n$, are the rows of the matrix $A_\kappa$\footnote{We implicitly assume the non-trivial case $A_\kappa\neq 0_\ntn$ since if $A_\kappa$ is the zero matrix, then one can select any $\tilde{w}_\kappa^{(\ell)}$ such that $|\Wt_\kappa|\neq 0$ and $\|\tilde{g} - g\|_{\L_{\infty}(\Omega)}=0$.}. By noticing that for $x\in \Omega$ we have 
\begin{equation}
\left\| (\tilde{W}_\kappa - W_\kappa  ) x \right\|  \leq \|x\|_{\L_{\infty}(\Omega)} \sum_{\ell = 1}^{r_\kappa} \| \tilde{w}_\kappa^{(\ell)}\|,
\end{equation}
and by looking at the $p$-th component of the difference $\tilde g - g $, by inequality \eqref{eq:ineq_wkappa_proof}, for $x\in\Omega$, we have
\begin{align}
   \left| \tilde{g}^{(p)}(x) - g^{(p)}(x) \right| & \leq L_\sigma \, \|x\|_  {\L_{\infty}(\Omega)} \sum_{\kappa \in K}\left( \|a_\kappa^{(p)}\| \sum_{\ell = 1}^{r_\kappa}  \|w_\kappa^{(\ell)}\|\right) \nonumber \\
    &\leq \sum_{\kappa \in K} \frac{\tilde{\varepsilon}}{\tilde{n}\sqrt{n}} \leq \frac{\tilde{\varepsilon}}{\sqrt{n}} \;,
\end{align}
from which we obtain $\|\tilde{g} - g\|_{\L_{\infty}(\Omega)} \leq \tilde\varepsilon$.
\end{proof}
\subsection{Proof of Theorem \ref{th:UAP_hamiltonian}}\label{app:th_main}
We prove the result by showing that the function $\varphi: \xi \mapsto q_N$ can be written in the form \eqref{eq:Cybenko_sum_matrix}, and thus, satisfying Proposition \ref{prop:UAPmaps}, it has the UAP on $\Omega$.
In fact, by restricting the parameter space as follows 
\begin{equation}\label{eq:parametrization_structure}
\begin{aligned}
J_j &= \begin{bmatrix} 0_\ntn & -X \\ X & 0_\ntn 
\end{bmatrix} \qquad  W_j = \begin{bmatrix} \tilde{W}_j & 0_\ntn \\ 0_\ntn & 0_\ntn
\end{bmatrix}, \\  
b_j &= \begin{bmatrix} \bt_j \\ 0_n  \end{bmatrix} \qquad\qquad\quad\quad \ \  \eta_j= \begin{bmatrix} 0_n \\ -\etat_j \end{bmatrix},
\end{aligned}
\end{equation}
where $X \in \Real^{n\times n}$, $\tilde{W}_j : \Realp \to \Real^{n\times n}$, $\bt_j: \Realp \to \Rn$, $\etat_j: \Realp \to \Rn$, one can write \eqref{eq:Hamiltonian_dynamics_SIE} as
\begin{equation}\label{eq:Hamiltonian_dynamics_SIE_proof_1}
\begin{aligned}
{\left[\begin{array}{c}
{p}_{j+1} \\
{q}_{j+1}
\end{array}\right] } & = \left[\begin{array}{c}
{p}_j  \\
{q}_j 
\end{array}\right]  + h \left[\begin{array}{c}
  X^\top \etat_j  \\
X \Wt_{j}^\top \sigma(\Wt_{j}p_{j+1} + \bt_j )
\end{array}\right]  \\
& = \left[\begin{array}{c}
{p}_j  \\
{q}_j 
\end{array}\right]  + \left[\begin{array}{c}
 \gammat_j  \\
\At_j \sigma(\Wt_{j}p_{j+1} + \bt_j  )
\end{array}\right] 
\end{aligned}
\end{equation}
for $j = 0, 1, \cdots, N-1$, where $\gammat_j = h X^\top \etat_j$, and $\At_j = h X \Wt_{j}^\top$, respectively. 
From the initial condition $(p_0,q_0) = (\xi,0_n)$,  $\xi \in \Omega$, and by substituting the expression of $p_{j+1}$ into the second equation of \eqref{eq:Hamiltonian_dynamics_SIE_proof_1} we have that
\begin{align}\label{eq:def_varphi}
    q_{N} &= \sum_{j = 0}^{N-1} \At_j \sigma(\tilde{W}_{j}( p_j + \gammat_j) + \bt_j ) \nonumber \\
    & = \sum_{j = 0}^{N-1} \At_j \sigma(\tilde{W}_{j} \xi + \dt_j) =: \varphi(\xi) \;,
\end{align}
where $\dt_j = \Wt_j \rt_j + \bt_j $ with $\rt_j=\rt_{j-1} + \gammat_j$ and $\rt_0=\gammat_0$.
Notice that, because of Lemma \ref{lem:W_full_rank}, we can assume, without loss of generality, that $\Wt_j$ in \eqref{eq:def_varphi} are full-rank.  
Consequently, one can freely choose $\At_j$  
by setting  $X = \frac{1}{h}\tilde{A}_j\tilde{W}_j^{-\top}$ for all $j=0,\dots,N-1$, while $\dt_j$ is free by construction due to the parameter $\bt_j$. 
Thus, the map \eqref{eq:def_varphi} 
has the UAP on $\Omega$ (Proposition \ref{prop:UAPmaps}),
\emph{i.e.} $\| \varphi(\xi) - g \|_{\L_\infty(\Omega)} \leq \tilde{\varepsilon}$, with $g$ in \eqref{eq:Cybenko_sum_matrix}. 
\hfill \QEDclosed

Note that the zero patterns of matrices, \emph{i.e.} $W, \eta, b$ in \eqref{eq:parametrization_structure} is only assumed for proving Theorem \ref{th:UAP_hamiltonian}.
However, since using more parameters\footnote{We recall that $J_j$ should keep the sparsity structure \eqref{eq:parametrization_structure} to maintain the non-vanishing gradients property (see \cite[Theorem 2]{galimberti2023hamiltonian}).} in \eqref{eq:parametrization_structure} cannot compromise UAPs, the structure of the weight matrices in \eqref{eq:parametrization_structure} is never used in practice.  
\subsection{Proof of Corollary \ref{cor:problem2}}\label{app_corollary}
In \cite[Proposition 3.8]{li2022deep} it is shown that there exists a continuous function $\psi(\xi)=\sum_{i=1}^N z_i \psi_i(\xi)$ for $\xi \in \Omega$,
where $z_i \in h^{-1}\left(F_i\right)$, with $\{F_i\}_{i=1}^N$ a partition of $f(\Omega)$, and continuous functions $\psi_i: \Omega \rightarrow[0,1]$ such that $\psi_i=1$ on  $A_i$ and $\psi_i=0$ on $\cup_{j \neq i} A_j$.
The sets $A_i\subset \Omega_i$, with $\{\Omega_i\}_{i=1}^N$ a partition of $\Omega$, such that $h \circ \psi$ has the UAP on $\Omega$ (provided that the desired function $f$ is such that $f\left(\Omega \right) \subseteq h\left(\mathbb{R}^n\right)$). Now, take $\psi$ such that $\left\|f-h \circ \psi\right\|_{\L_{\infty}(\Omega)} \leq \varepsilon\slash 2$ and, by Theorem \ref{th:UAP_hamiltonian}, take $\varphi : \xi \mapsto q_N$ such that $\left\|\psi - \varphi \right\|_{\L_{\infty}(\Omega)}\leq \varepsilon\slash (2 L_h).$
Then, for any $\xi \in \Omega$ we have
\begin{equation*}
\begin{aligned}
\|f-h \circ \varphi (\xi)\| & \leq\left\|f(\xi)-h \circ \psi(\xi) \right\|\\ 
&\qquad+\left\|h(\xi) \circ \psi(\xi) - h \circ \varphi(\xi)\right\| \\
& \leq \frac{\varepsilon}{2}+L_h\left\|\psi(\xi)-\varphi(\xi)\right\| \leq \varepsilon\;, 
\end{aligned} 
\end{equation*}
and the proof is completed. \hfill \QEDclosed
\subsection{Proof of Proposition \ref{prop:approx}}\label{app:approx}
 The proof follows from \cite[Theorem 1]{barron1994approximation} by noting that the function $\varphi$ in \eqref{eq:def_varphi} is the NN considered in \cite{barron1994approximation}, by selecting the probability measure $\tilde{\lambda}(\cdot) := \frac{1}{2^n} \lambda(\cdot)$ where $\lambda$ is the Lebesgue measure on $\Omega$, and by recalling the norm inequality $\|\cdot \|_\infty\leq \|\cdot \|_2$.\hfill \QEDclosed


\bibliography{bibliography.bib}
\bibliographystyle{unsrt}
\end{document}